%% file: ijcai22.tex
\documentclass{article}
\usepackage{ijcai22}

\usepackage{times}
\usepackage{soul}
\usepackage{url}
\PassOptionsToPackage{hyphens}{url}\usepackage[hidelinks]{hyperref}
\usepackage[utf8]{inputenc}
\usepackage[small]{caption}
\usepackage{graphicx}
\usepackage{amsmath}
\usepackage{amsthm}
\usepackage{booktabs}
\usepackage{algorithm}
\usepackage{algorithmic}
\usepackage{bbm}
\usepackage{latexsym}

\usepackage{tikz}
\usepackage{amsfonts}
\usepackage{csvsimple}
\usepackage{pgfplots}
\usepackage{subcaption}
\pgfplotsset{compat = newest}
\DeclareMathOperator*{\argmax}{arg\,max}

\usepackage{array}
\usepackage{colortbl}
\usepackage{multirow}
\usepackage{multicol}
\usepackage{xfp}
\usepackage{hhline}
\usepackage{siunitx}

\title{Blind Knowledge Distillation for Robust Image Classification}

\author{
Timo Kaiser\and
Lukas Ehmann\and
Christoph Reinders\And
Bodo Rosenhahn
\affiliations
Institute for Information Processing, Leibniz University Hannover
\emails
\{kaiser, ehmannlu, reinders, rosenhahn\}@tnt.uni-hannover.de
}

\begin{document}

\maketitle

\input{0_abstract}

\input{1_introduction}
\input{2_related_work}
\input{3_0_preliminaries}
\input{3_method}

\input{4_experiments}
\input{5_conclusion}
\input{6_acknowledgements}

\bibliographystyle{named}
\bibliography{ijcai22}

\end{document}

%% file: 0_abstract.tex
\begin{abstract}
Optimizing neural networks with noisy labels is a challenging task, especially if the label set contains real-world noise. Networks tend to generalize to reasonable patterns in the early training stages and overfit to specific details of noisy samples in the latter ones. We introduce \textit{Blind Knowledge Distillation} - a novel teacher-student approach for learning with noisy labels by masking the ground truth related teacher output to filter out potentially corrupted `knowledge' and to estimate the tipping point from generalizing to overfitting. Based on this, we enable the estimation of noise in the training data with Otsu's algorithm. With this estimation, we train the network with a modified weighted cross-entropy loss function. We show in our experiments that \textit{Blind Knowledge Distillation} detects overfitting effectively during training and improves the detection of clean and noisy labels on the recently published CIFAR-N dataset. Code is available at GitHub\footnote{\href{https://github.com/TimoK93/blind_knowledge_distillation}{https://github.com/TimoK93/blind\_knowledge\_distillation}}.  
\end{abstract}

%% file: 1_introduction.tex
\section{Introduction}
\input{plot_distribution}

Learning with noisy labels is a challenging task in image classification. It is well known that label noise leads to heavy performance drops with standard classification methods~\cite{9729424}. The goal of learning with noisy labels is therefore to train a classification model with labelled training images and achieve high classification performance on unseen test images, even if the labels for training are noisy and corrupted. Labels are noisy because humans are naturally unable to classify images perfectly due to ambiguous images, individual human bias, pressure of time, or various other reasons. Many modern methods~\cite{Liu_2022_CVPR,deepconv} are trained on large and potentially noisy datasets and thus it is an interest of the community to make classification robust against noisy labels.  

To evaluate the robustness of methods for learning with noisy labels, clean image datasets like \textit{CIFAR}~\cite{krizhevsky2009learning}, \textit{Clothing1M}~\cite{xiao2015learning}, or \textit{WebVision}~\cite{li2017webvision} are synthetically corrupted by randomly flipping label annotations either symmetrically without constraints or asymmetrically with predefined rules to mimic realistic label noise. However, \citeauthor{wei2022learning}~\shortcite{wei2022learning} shows that synthetic label noise has different behaviour compared to real-world label noise and is thus not an ideal choice to evaluate robust learning. To close this gap, \citeauthor{wei2022learning} have made great efforts and presented \textit{CIFAR-N} with multiple newly annotated ground truth labels for \textit{CIFAR} with human-induced label noise. With these new annotations, robust learning can be evaluated more realistically. 

In this paper, we introduce a novel method to detect the beginning of overfitting on sample details during training, that is usually roughly estimated as in \cite{Li2020DivideMix:}, and present a simple but effective method to detect most likely corrupted labels. 
Our method is inspired by \textit{Knowledge Distillation}~\cite{hinton2015distilling} for neural networks which extracts `knowledge' from a teacher network to train a student network. Differently than usual, our student network is just trained with a subset of the teachers `knowledge'. Specifically, it does not `see' the `knowledge' about the given and potentially corrupted ground truth labels by utilizing the teachers ground truth complementary logits. Therefore we call it \textit{Blind Knowledge Distillation}. Based on the detected noisy labels, we propose a simple but effective loss-correction method to train the teacher model robustly with label noise.
We perform extensive experiments on CIFAR-10N and the results show that \textit{Blind Knowledge Distillation}
\begin{itemize}
    \item successfully estimates the tipping point from fitting to general patterns to (over)fitting to sample details,
    \item is an effective method to estimate the likelihood of labels being noisy,
    \item and improves the classification accuracy while training with high noise levels.
\end{itemize}

%% file: plot_distribution.tex
\begin{figure}[t]

\begin{tikzpicture}
\begin{axis}[
    scaled ticks=false, 
    log ticks with fixed point,
    tick label style={/pgf/number format/fixed},
    xmin = 0, xmax = 1.0,
    ymin = 0, ymax = 0.4,
    xtick distance = 0.25,
    ytick distance = 0.1,
    grid = both,
    minor tick num = 2,
    major grid style = {lightgray},
    minor grid style = {lightgray!25},
    width = 1\linewidth,
    height = 0.9\linewidth,
    xlabel = {$P(y=\overline{y}|x)$},
    ylabel = {$n$},
    legend style = {
      at={(0.75,0.99)},
      legend columns=1},
]

\addplot[
    ybar, 
    red,
    fill=red,
    bar width=0.01,
    x=0.02,
    solid,
] table[x=x, y=f ] {distribution.txt};

\addplot[
    ybar, 
    green,
    fill=green,
    bar width=0.01,
    x=0.02,
    bar shift=0.01, 
    solid,
] table[x=x, y=p ] {distribution.txt};

\addplot[
    thick,
    blue,
    solid,
] table[x=x, y=g_low ] {gauss.txt};

\addplot[
    thick,
    blue,
    solid,
] table[x=x, y=g_up ] {gauss.txt};
\legend{Noisy Labels, Clean Labels, Otsu Distributions}

\addplot +[mark=none,black, thick, dashed] coordinates {(0.1282, 0) (0.1282, 1.0)};
\addplot +[mark=none,black, thick, dashed] coordinates {(0.52, 0) (0.52, 1.0)};
\addplot +[mark=none,black, thick, dashed] coordinates {(0.8612, 0) (0.8612, 1.0)};
\node[draw,align=left] at (0.19, 0.25) {$\mu_1$};
\node[draw,align=left] at (0.565, 0.25) {$s$};
\node[draw,align=left] at (0.92, 0.25) {$\mu_2$};
\end{axis}
\end{tikzpicture}

\caption{Distribution of ground truth label related \mbox{probabilities $P_A(y=\overline{y}|x)$} at beginning of overfitting (tipping point) and the resulting gaussian distributions after Otsu's algorithm for the dataset \textit{Worst}. Red bars show the normalized distribution of noisy labels and green bars of clean labels, respectively. Note that the gaussian distributions (blue) are scaled for visualization purposes. Our presented Blind Knowledge Distillation enables an adaptive noise estimation via the thresholds $\mu_1$, $s$, and $\mu_2$ and a robust learning with noisy labels.}
\label{fig:distribution}
\end{figure}
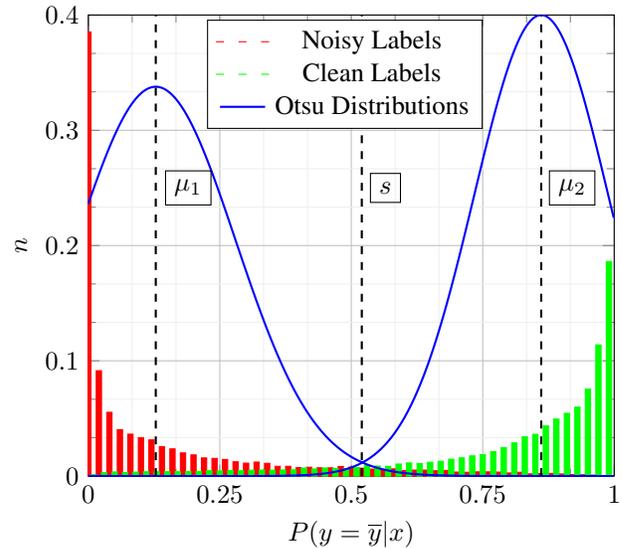

%% file: 2_related_work.tex
\section{Related Work}
Methods in the field of robust learning to tackle noisy labels can be divided into label correction, loss correction, and refined strategies~\cite{9729424,Wang_2019_ICCV}. In this section, we contextualize the latest methods of the \textit{CIFAR-N} leaderboard based on the aforementioned categories.

Label correction is an approach in which the given ground truth labels are dynamically changed during optimization to obtain labels of higher quality. \textit{SOP}~\cite{https://doi.org/10.48550/arxiv.2202.14026} performs label correction by optimizing the ground truth labels with Stochastic Gradient Descent (SGD) w.r.t.~the classification loss. It alternates between the update of model weights and the update of additional soft-label weights.

Another approach is loss correction which is usually applied by weighting the loss term or adding a new loss for each sample in the training dataset. The methods \textit{CORES}~\cite{cheng2021learning} and \textit{ELR}~\cite{NEURIPS2020_ea89621b} add a regularization term to the standard cross-entropy (CE) loss to penalize likely corrupted labels. \textit{PeerLoss}~\cite{liu2020peer} introduces and minimizes peer loss functions between randomly selected samples. \textit{CAL}~\cite{zhu2021second} extends this approach and estimates the covariances between noise rates and their bayes optimal label.

The last category tackles noisy labels by using refined strategies. \textit{CoTeaching}~\cite{NEURIPS2018_a19744e2} trains a neural network with samples with high confident predictions of a second network, and vice versa. \textit{DivideMix}~\cite{Li2020DivideMix:} and \textit{PES}~\cite{NEURIPS2021_cc7e2b87} split the dataset in clean and corrupted subsets and apply semi-supervised learning methods. In detail, \textit{DivideMix} trains two independent neural networks and splits the set of one network based on the predictions of the other network to avoid confirmation bias. In contrast to this, \textit{PES} applies early stopping of the optimization to every network layer independently, instead of applying it to the whole network simultaneously, as usual.

Our method combines a refined strategy to detect most likely corrupted labels in the first stage and performs loss-correction in the second stage while incorporating the estimation of likely corrupted labels. While other methods manually define warm-up epochs, we adapt to the dataset and estimate the optimal stopping point for the standard CE training. Instead of applying extensive semi-supervised augmentation methods, we apply a simple sample dependent loss correction.  

%% file: 3_0_preliminaries.tex
\section{Preliminaries}
Given a set of annotated image samples $X$ and a set of classes $C$, the task of image classification is to assign every sample $x$ from $(x,\overline{y})\in X$ to the correct class label $y=c\in C$ without prior knowledge of the correct class label $y$ and a potentially noisy annotation $\overline{y}$. Modern methods use neural networks $f(\Phi, x)$ to estimate the probability distribution  $P(y=c|x)$ for every class $c$~\cite{Liu_2022_CVPR,Li2020DivideMix:,https://doi.org/10.48550/arxiv.2202.14026,cheng2021learning,he2016deep}, in which $\Phi$ denotes a set of trainable network parameters. More specifically, neural networks predict a logit vector $\vec{l}\in \mathbb{R}^{|C|}$ with a logit $l_{c}$ for every class and transform it into probabilities with the softmax function 
\begin{equation}
\label{equ:softmax}
    P(y=c|x) = \frac{e^{l_c}}{\sum_{i\in C}e^{l_i}}.
\end{equation}
Finally, the class $c$ with the highest probability $P(y=c|x)$ is assumed to be the correct label $y$.

The task is to define the network architecture of $f$ and the training strategy to optimize $\Phi$, so that $f(\Phi, x)$ predicts a satisfying distribution $P(y=c|x)$ in which the correct class has the highest probability. Most methods optimize $\Phi$ with large manually annotated image datasets and minimize the categorical cross entropy (CE) loss objective
\begin{equation}
\label{equ:ce}
    L_{\text{CE}}= \frac{1}{|X|} \sum_{\substack{(x,\overline{y})\in X}}  -\log \big(P(y=\overline{y}|x)\big)
\end{equation}
or one of its derivatives.

Extending the task of image classification, the challenging task of learning with noisy labels addresses the problem that the given ground truth labels $\overline{y}$ could be noisy and not the true labels $\overline{y}\neq y$. False ground truth labels dramatically impede the optimization of $\Phi$. Thus, the goal is to train classifiers with an accuracy that is comparable to classifiers that would be optimized with clean labels $\overline{y}=y$. A second goal is to identify noisy labels $\overline{y}\neq y$ in the dataset. 

The approach proposed in this paper addresses both tasks. Note that the method is iteratively trained with random sampled batches $X'\subset X$. We keep the notation of $X$ in the next sections for simplicity, e.g.~in Eq.~\eqref{equ:stud_loss}.    

%% file: 3_method.tex
\section{Method}
\begin{figure*}[t]
    \centering
    \includegraphics[width=1\textwidth]{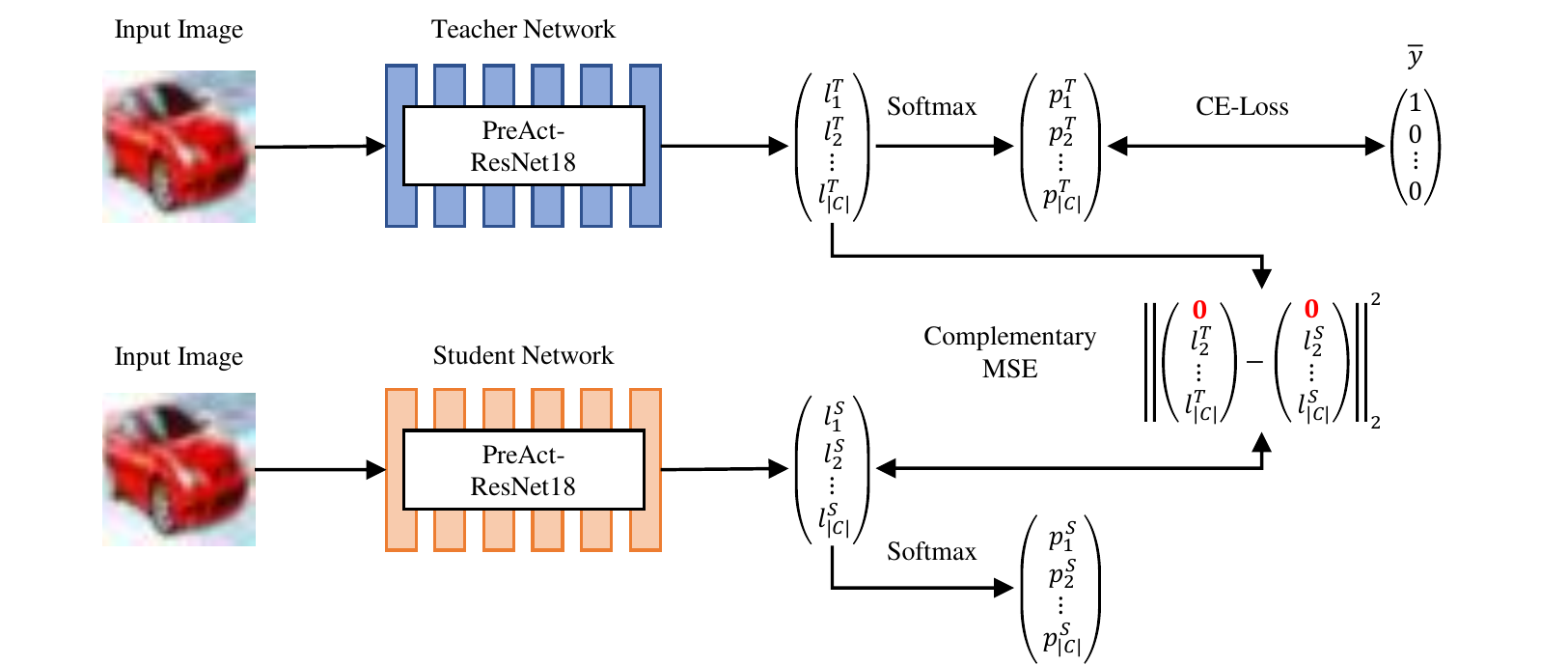}
    \caption{Our proposed \textit{Blind Knowledge Distillation} framework. The teacher and student network share the same topology but have different weights. While the teacher network is trained by optimizing the standard CE loss, the student network is trained with the ground truth complementary logits of the teacher network and the mean squared error loss. The teacher network predicts the class membership \mbox{probabilities $P_T(y=c|x)$ and} the student network predicts the probabilities $P_S(y=c|x)$.}
    \label{fig:framework}
\end{figure*}

To enlarge the robustness of neural networks against label noise and to detect noisy labels, we present a novel training strategy to estimate the likelihood of every label being noisy and apply a weighted loss based on this. 

First, we adapt the student-teacher architecture~\cite{gou2021knowledge} and introduce \textit{Blind Knowledge Distillation} to extract generalized patterns from the data. Then, we present a method to detect the beginning of overfitting with the student network and enable the detection of noisy labels by estimating four confidence levels of being noisy. Finally, we optimize $\Phi$ with a robust training strategy to train the final classifier. All three steps are described in the following sections. 

\subsection{Blind Knowledge Distillation}
Neural networks with large number of parameters $\Phi$ can memorize the training examples, so that $P(y=\overline{y}|x)\approx 1$ for every sample $x$ in the training set $X$. Also, neural networks adapt simple patterns during the early optimization epochs and overfit to specific image details in the latter ones. 
As shown in~\cite{NEURIPS2020_ea89621b}, valid patterns are learned by maximizing the logit $l_{y=\overline{y}}$ for clean samples in the first training stages. Subsequently to this early generalization, maximizing the logits $l_{y\neq \overline{y}}$ of corrupted samples degrades the classification accuracy. More important for this method is the phenomenon that the ground truth complementary logits $l_{c\neq \overline{y}}$ are also minimized in the latter stages. 

To avoid the maximization of $l_{y\neq\overline{y}}$ for noisy labels, we create a new student-teacher architecture, in which the student only learns generalized patterns. In the student-teacher architecture, a student model is trained with the output of a teacher model. This method is called \textit{Knowledge Distillation}~\cite{hinton2015distilling} and transfers the patterns that are encoded in the teacher model to the student. Model bias or wrong patterns from the teacher can also be transferred. To avoid this undesired transfer, we introduce the ground truth annotation complementary `knowledge' by removing all information that is immediately connected to a potentially corrupted ground truth label $\overline{y}$. 

Unlike the usual knowledge distillation architecture, our models share the same topology $f(\Phi, x)$ but have different weights $\Phi_T$ (teacher) and $\Phi_S$ (student). The teacher model $f(\Phi_T, x)$ is trained with the standard CE \mbox{loss (Eq.~\eqref{equ:ce}).} The student model is trained with the unlabelled ground truth complementary logits $l^T_{c\neq\overline{y}}$ derived from the teacher model and an extended but simple mean squared error loss
\begin{equation}
\label{equ:stud_loss}
L_{\text{Stud}} = \frac{1}{|X|} \sum_{\substack{(x,\overline{y})\in X}} \frac{1}{|C| - 1} \sum_{\substack{c\in C\\c\neq \overline{y}}} (l^T_{c}-l^S_{c})^2
\end{equation}
in which $l^S_{c}$ denotes the logits of the student model. This loss function transfers the generalized patterns by imitating the output of the teacher model but without taking the potentially corrupted ground truth label $\overline{y}$ into account. The training architecture is visualized in Fig.~\ref{fig:framework}.

Since high valued complementary logits $l_{c\neq \overline{y}}$ are minimized in the latter optimization stages after general features are learned, also the resulting probabilities after \mbox{softmax (Eq.~\eqref{equ:softmax})} converge to a uniform distribution. Thus, we can identify approximately the training epoch, in which the neural network starts overfitting to specific sample details by monitoring the mean maximal probability
\begin{equation}
\label{equ:mean_max}
\hat{p}_{\text{max}} = \frac{1}{|X|}\sum_{x\in X} \max_{c\in C}\big(P(y=c|x)\big)
\end{equation}
of the students network. During training in epoch $i$, the \textit{fitting-epoch} in which $\hat{p}^i_{\text{max}}$ is maximal can be certainly identified online with a delay of $k$ epochs by checking if $\hat{p}^{i-k}_{\text{max}}$ is the maximum of the last $2k+1$ epochs.     

Furthermore, it shows that the student model has the ability to classify images comparable to the teacher model before the detail fitting starts.  
Thus, we modify our final classification probability by combining the teacher's \mbox{prediction $P_T(y=c|x)$} and the student's prediction $P_S(y=c|x)$ to the \textit{agreement} probability
\begin{equation}
\label{equ:agree}
P_A(y=c|x) = \frac{P_T(y=c|x)\cdot P_S(y=c|x)}{\sum_{i\in C} P_T(y=i|x)\cdot P_S(y=i|x)}. 
\end{equation}
$P_A$ enables the noise estimation in the dataset described in the following. The classification accuracy $P_T$, $P_S$, and $P_A$ are elaborated more in detail in the experiments (see Sec.~\ref{sec:knowledge}). 

\subsection{Adaptive Noise Estimation}
The knowledge about the presence of noise can be used to apply loss correction. Unfortunately, this knowledge is not given, so we estimate the probability of a data sample to be noisy.
We split the dataset into four subsets based on Otsu's algorithm~\cite{otsu1979threshold}, in which the membership to a subset indicates the likelihood of being noisy. Given a set of data samples $(x, \overline{y}) \in X$ with their corresponding agreement probability $P_A(y=\overline{y}|x)$, the first step is to find a threshold $s$ that splits $X$ into two distributions $X_{1}=\{x\in X|P_A(y=\overline{y}|x) \leq s \}$ and $X_\text{2}=\{x\in X|P_A(y=\overline{y}|x) > s  \}$, in which $X_{1}$ contains images with likely noisy and $X_\text{2}$ images with likely clean labels. To find $s$, we assume that $X_{1}$ and $X_{2}$ can be approximated by two gaussian distributions $(\mu_1,\sigma_1)$ and $(\mu_2,\sigma_2)$. The optimal threshold $s$ maximizes the objective 
\begin{equation}
    Q(s)=\frac{n_1(s)\big( \mu_1(s) -\mu \big)^2 + n_2(s)\big( \mu_2(s) -\mu \big)^2}{n_1(s)\sigma_1(s)^2 + n_2(s)\sigma_2(s)^2},
\end{equation}
where $n_1(s)$ and $n_2(s)$ denote the cardinality of $X_1$ and $X_2$ depending on $s$, and $\mu$ is the mean probability $P_A(y=\overline{y}|x)$ of all samples in $X$. The optimal $s$ minimizes the inter-class variance and can be found by calculating $Q(s)$ for all $s$ with a reasonable step size $\Delta s = 0.001$. 

Using Otsu's algorithm, we preserve a threshold $s$ to split the data into noisy and clean samples, and furthermore thresholds $\mu_1$ and $\mu_2$ to subdivide the subsets into more fine-grained subsets. A finer distinction w.r.t. the likelihood of being noisy allows a more precise weighting of the samples in the following steps. Depending on the requirements of the application, the task of label noise detection can be solved by classifying a sample $x$ by comparing $P_A(y=\overline{y}|x)$ with one of the thresholds $s$, $\mu_1$, and $\mu_2$. While using $\mu_1$ is more liberal to classifying noisy labels into the clean dataset than $s$, $\mu_2$ is more conservative. A visualization of a distribution of $P_A(y=\overline{y}|x)$ and the estimated noise is shown in Fig.~\ref{fig:distribution}.

\subsection{Robust Optimization}
\label{sec:robust_optimization}
After splitting up the dataset into potentially clean and corrupted data, we use simple robust training techniques to train the final classification model. Based on the idea of label smoothing~\cite{7780677}, we extend the CE \mbox{loss (Eq.~\eqref{equ:ce})} and combine the ground truth label with the student's prediction and a sample dependent $\alpha_x$: 
\begin{multline}
\label{equ:ce_boot}
    L_{\text{Robust}}= -\frac{1}{|X|} \sum_{(x,\overline{y})\in X} \sum_{c\in C}  \mathcal{S}(\beta^c_x)\log\big(P_T(y=c|x)\big) \\
    \text{with}\ \beta^c_x =  (1-\alpha_x)\mathbbm{1}[c=\overline{y}] + \alpha_x P_S(y=c|x) 
\end{multline}
and \textit{Sharpening} $\mathcal{S}$ that is explained later.

While the teacher network is trained by $L_{\text{Robust}}$, the student network is still trained with the student loss \mbox{$L_{\text{Stud}}$ (Eq.~\eqref{equ:stud_loss})}. As larger as the instance dependent $\alpha_x$ gets, the less the ground truth of a sample $x$ is trusted. 
We adapt $\alpha_x$ for every sample individually, depending on the cluster membership after Otsu. We define four fixed alpha values with $\alpha^1<\alpha^2<\alpha^3<\alpha^4$ where $\alpha^1$ gets assigned to samples with $P_A(y=\overline{y}|x)\geq\mu_2$, $\alpha^2$ to samples with $\mu_2>P_A(y=\overline{y}|x)\geq s$, $\alpha^3$ to samples with $s>P_A(y=\overline{y}|x)\geq \mu_1$, and $\alpha^4$ otherwise.

Since a larger $\alpha_x$ enlarges the entropy in the objective, we use a modified \textit{Sharpening} method
\begin{equation}
\label{eq:sharpening}
\mathcal{S}(\beta^c_x) = \frac{(\beta^c_x)^{1+\alpha_x}}{\sum_{i\in C}{(\beta^i_x)^{1+\alpha_x}}} 
\end{equation}
as used by~\cite{Li2020DivideMix:} to minimize the entropy. The Sharpening function is applied stronger for insecure samples by reusing the above mentioned alpha.

%% file: 4_experiments.tex
\section{Experiments}
\input{table_classification}

\input{table_detection}
\input{plot_fig1}

\input{plot_accuracy_worse}

We perform several experiments to evaluate our proposed method. The experimental setup and the used metrics are explained first. Then we present evaluation metrics on the recently released dataset \textit{CIFAR-10N}~\cite{wei2022learning} and show details and observations of our core method \textit{Blind Knowledge Distillation}.

\subsection{Experimental Setup}
We evaluate our method on the noise levels provided in the CIFAR-10N dataset.
To be comparable to other methods, we utilize the same model setup as used in~\cite{Li2020DivideMix:}. We use a 18-layer PreAct ResNet~\cite{he2016identity} and  \textit{Stochastic Gradient Descent} with momentum of $0.9$ and weight decay of $0.0005$ as optimizer. The networks are trained for $300$ epochs beginning with a learning rate of $0.02$ and reduce it to $0.002$ after $150$ epochs. We train the network with randomly sampled batches of $128$ image samples. In the first stage, the teacher network optimizes the standard CE loss (Eq.~\eqref{equ:ce}) until the detection of the tipping point induces the start of the second stage, in which the teacher network optimizes the modified loss (Eq.~\eqref{equ:ce_boot}). The hyperparameters introduced by our method are set to $\alpha^1=0.3$, $\alpha^2=0.45$, $\alpha^3=0.55$, $\alpha^4=0.7$. For the noisy detection task, we use the probabilities $P_A$ and the threshold $\mu_1$ after Otsu. We repeated the experiments at least five times with random seeds and report the averaged metrics.

The method proposed in this paper is evaluated on \textit{CIFAR-10N}~\cite{wei2022learning}. CIFAR-10N manually re-labelled the CIFAR-10~\cite{krizhevsky2009learning} by multiple humans to investigate the impact of realistic label noise compared to synthetically induced ones. The dataset contains 50K training images and 10K test images with a size of $32\times32$. For the training set, there are five label sets with realistic human label noise with a ratio of approx. $9\%$, $17\%$, $18\%$, $18\%$, and $40\%$ label noise. In the same order of the noise ratios, we denote them as \textit{Aggre}, \textit{Rand1}, \textit{Rand2}, \textit{Rand3}, and \textit{Worst} in our experiments. 

The tasks for the dataset are twofold: First, the classifier should be trained robust to achieve a high test accuracy even with high noisy rates and second, noisy labels in the training data should be detected and marked as noisy. The metrics to evaluate the tasks are given in the next section.

\subsection{Metrics}
We evaluate the performance of image classification with the commonly used \textit{Accuracy} (Acc) metric. It measures the classification accuracy on the test dataset $X_\text{Test}$ using the ratio of correct classified test samples compared to the dataset size:
\begin{equation}
\text{Acc}=\frac{\sum_{(x, \overline{y}) \in X_\text{Test}} \mathbbm{1}\big[\argmax_{c\in C}\big(P(y=c|x)\big) = \overline{y}\big]}{|X_\text{Test}|}    
\end{equation}

The task of noisy label detection is evaluated with the well-known \mbox{$F_1$-score}, \textit{Precision} (Pr), and \textit{Recall} (Re) metrics, in which \textit{Precision} decreases if clean labels are classified as noisy and \textit{Recall} decreases if noisy labels are classified as clean. The $F_1$-score harmonizes both aspects. With the subsets of true ($X_{\text{Noise}}\subset X$) and predicted ($X'_{\text{Noise}}\subset X$) noisy labels from the training set $X$, the metrics are defined as:
\begin{equation}
\text{Pr}=\frac{\sum_{x \in X'_{\text{Noise}}} \mathbbm{1}\big[x\in X_{\text{Noise}}]}{|X'_{\text{Noise}}|},    
\end{equation}
\begin{equation}
\text{Re}=\frac{\sum_{x \in X_{\text{Noise}}} \mathbbm{1}\big[x\in X'_{\text{Noise}}]}{|X_{\text{Noise}}|},   
\end{equation}
\begin{equation}
\text{and} \quad F_1=\frac{2}{\text{Pr}^{-1}+\text{Re}^{-1}}.    
\end{equation}

\subsection{CIFAR-10N}
This section elaborates the results for the tasks of robust training and noise detection.

\paragraph{Robust Training}
We compare our results to the latest six state-of-the-art methods and the standard CE baseline on the CIFAR-10N Leaderboard in Tab.~\ref{tab:classification}. Our method achieves the performance to be listed on the new sixth position of the leaderboard outperforming \textit{CAL} and standard \textit{ELR}. We want to mention that \textit{ELR+} and \textit{DivideMix} apply multiple models and high performance semi-supervised strategies such as \textit{MixMatch}~\cite{NEURIPS2019_1cd138d0}. 

\paragraph{Noise Detection}
The detection performance is shown in Tab.~\ref{tab:detection}. We present $F_1$, precision, and recall for all five noise levels in CIFAR-10N (10 classes). The split to classify clean and corrupted labels is performed based on one of the probability sets $P_T$, $P_S$, and $P_A$ and the three threshold $s$, $\mu_1$, $\mu_2$ provided by Otsu's method. Intuitively, the precision is higher for the lower threshold $\mu_1$ and recall for the higher threshold $\mu_2$, respectively. The experiments show that the harmonized metric $F_1$ performs best for $\mu_1$. Thus, $\mu_1$ is used to solve the task of Noise detection. The combined probability $P_A$ performs better or on par w.r.t.~the $F_1$-score, confirming the improved classification accuracy that is also visible in Fig.~\ref{fig:accuracy_noise}. An exemplary distribution of ground truth label probabilities for clean and noisy labels with the subsequent split based on Otsu is shown for the \textit{Worst} dataset in Fig.~\ref{fig:distribution}.

\subsection{Blind Knowledge Distillation}
\label{sec:knowledge}
The core contribution of our method is \textit{Blind Knowledge Distillation}. This section analyzes its ability to detect (over)fitting on sample details and how it can be used to improve the classification accuracy. For the experiments in this section, we trained our student-teacher framework without detection of the tipping point and robust optimization (Sec.~\ref{sec:robust_optimization}) after generalization for 75 epochs to show the teacher's and student's learning behavior. 

Fig.~\ref{fig:pmax} shows the average maximal probability prediction of the student network over training time. The probability strongly increases in the first training epochs and degrades after a tipping point. It is notable that high noise rates decrease the absolute mean probability in general (see \textit{Worst}). Our explanation of this behavior is that classifiers are not able to clearly predict a class based on simple and generalized but ambiguous image patterns. Thus, the classifier produces multiple predictions $P(y=c|x)\gg 0$ during the first generalization stage. An example pattern could be the coarse shape which is often ambiguous, e.g.~for classes \textit{dog} and \textit{cat}. In the second training stage after the tipping point, the teacher network adapts detailed sample patterns to maximize $P(y=\overline{y}|x)$ which also minimizes $P(y\neq \overline{y}|x)$  

The test classification accuracy of the teacher network is shown in Fig.~\ref{fig:acc}. While early stopping of standard optimization is not important for clean datasets or low noise levels (\textit{Aggre}), fitting on sample details leads to overfitting and decreases the classification accuracy on high noise rates (\textit{Worst}). Therefore, the choice of an early stopping epoch is highly important. It shows that the tipping point from Fig.~\ref{equ:mean_max} is a good indicator to detect overfitting. It proposes an accurate estimation to stop optimizing on high noise levels without stopping too early on low noise levels. 

Using the tipping point in Fig.~\ref{fig:pmax} to split the data into potentially corrupted and clean subsets is intuitively, due to the beginning overfitting and decreasing classification accuracy. Fig.~\ref{fig:f1} shows the detection ability with the $F_1$-score if splitting the dataset based on Otsu's algorithm and $P_A$ at every epoch. Similar to Fig.~\ref{fig:acc}, the tipping point gives a guess for a suitable epoch to split the dataset. The estimation for high noise rates is sufficient, especially due to the decreasing $F_1$-score after the tipping point. Since low noise rates does not seem to affect the $F_1$ negatively in latter training stages, the tipping point estimate leads to a slightly too early splitting epoch.

An interesting insight about the teacher and student classification accuracy $P_T$ and $P_S$ is shown in Fig.~\ref{fig:accuracy_worse_all} on a high noise level. While the teachers accuracy decreases during overfitting, the students accuracy persists. We claim that using the complementary student loss from Eq.~\eqref{equ:stud_loss} prevents the student from fitting to misleading image details by removing the ground truth related logits $l_{c=\overline{y}}$. Also interesting is that the combined probability $P_A$ suits as the overall best probability for classification. While the combined probability is quite similar to the teachers probability $P_A\approx P_T$ during the early stage, it converges to the students accuracy $P_A\approx P_S$ in the latter ones. Near the tipping point, it outperforms both.       

Overall, \textit{Blind Knowledge Distillation} is a better choice to automatically detect overfitting rather than to stop training after predefined and fixed periods (e.g.~in~\cite{Li2020DivideMix:}). Combining $P_T$ and $P_S$ to $P_A$ can be used to improve the overall classification accuracy.

%% file: table_classification.tex
\begin{table}[t]
    \centering
    \resizebox{\columnwidth}{!}{%
    \begin{tabular}{c|c c c c c}
    \toprule
    Acc [\%] & Aggre & Rand1 & Rand2 & Rand3 & Worst
    \\\midrule
    \textit{SOP} & \textbf{95.61} & \textbf{95.28} & \textbf{95.31} & \textbf{95.39} & \textbf{93.24}\\
    \textit{CORES} & 95.25 & 94.45 & 94.88 & 94.47 & 91.66\\
    \textit{DivideMix} & 95.01 & 95.16 & 95.23 & 95.21 & 92.56 \\
    \textit{ELR+} & 94.83 & 94.43 & 94.20 & 94.34 & 91.09 \\
    \textit{PES} & 94.66 & 95.06 & 95.19 & 95.22 & 92.68 \\
    \textit{ELR} & 92.38 & 91.46 & 91.61 & 91.41 & 83.58 \\
    \textit{CAL} & 91.97 & 90.93 & 90.75 & 90.74 & 85.36 \\
    CE & 87.77 & 85.02 & 86.14 & 85.16 & 77.69 \\\midrule
    Ours & 93.68 & 92.50 & 92.63 & 92.54 & 86.64 \\
    \bottomrule
    \end{tabular}}
    \caption{Classification accuracy of our method compared to standard CE-loss framework and state-of-the-art methods
    \textit{SOP}~\protect\cite{https://doi.org/10.48550/arxiv.2202.14026},
    \textit{CORES}~\protect\cite{cheng2021learning},
    \textit{DivideMix}~\protect\cite{Li2020DivideMix:},
    \textit{PES}~\protect\cite{NEURIPS2021_cc7e2b87},
    \textit{ELR}~\protect\cite{NEURIPS2020_ea89621b}, and
    \textit{CAL}~\protect\cite{zhu2021second}
    .}
\label{tab:classification}
\end{table}     

%% file: table_detection.tex
\begin{table*}[t]
    \centering
    \begin{tabular}{p{0.02\linewidth} p{0.02\linewidth} | c c c | c c c | c c c |c c c | c c c}
    \toprule
    \multicolumn{2}{c|}{\multirow{2}{*}{$[\%]$}} & \multicolumn{3}{c|}{Aggre} & \multicolumn{3}{c|}{Rand1} & \multicolumn{3}{c|}{Rand2} & \multicolumn{3}{c|}{Rand3} & \multicolumn{3}{c}{Worst} \\ 
    \multicolumn{2}{c|}{} & $F_1$ & Pr & Re & $F_1$ & Pr & Re & $F_1$ & Pr & Re & $F_1$ & Pr & Re & $F_1$ & Pr & Re \\
    \midrule
    
    \multirow{3}{*}{\rotatebox{90}{$P_T$\hspace{0pt}}}     
  & $\mu_1$ &  69.2 &  58.0 &  85.9&  \textbf{82.9} &  82.7 &  83.0&  \textbf{83.4} &  83.6 &  83.3&  \textbf{83.1} &  82.9 &  83.2&  75.0 &  \textbf{95.9} &  61.6 \\
  & $s$ &  46.6 &  30.7 &  97.1&  67.2 &  51.3 &  97.5&  67.9 &  52.1 &  97.6&  67.3 &  51.3 &  97.6&  85.9 &  79.6 &  93.3 \\
  & $\mu_2$ &  28.8 &  16.8 &  99.7&  44.8 &  28.8 &  99.8&  45.9 &  29.8 &  99.8&  45.1 &  29.1 &  99.8&  70.8 &  55.0 &  99.6  \\\midrule
   
    \multirow{3}{*}{\rotatebox{90}{$P_S$\hspace{0pt}}} 
  & $\mu_1$ &  50.9 &  35.7 &  88.4&  74.5 &  65.7 &  86.0&  75.8 &  67.9 &  85.9&  74.8 &  66.1 &  86.0&  75.8 &  92.3 &  64.4 \\
  & $s$ &  31.9 &  19.0 &  98.3&  55.0 &  38.2 &  98.2&  56.6 &  39.7 &  98.1&  55.4 &  38.6 &  98.5&  81.9 &  72.2 &  94.4 \\
  & $\mu_2$ &  22.3 &  12.5 &  \textbf{99.8}&  39.8 &  24.8 &  \textbf{99.9}&  41.3 &  26.0 &  \textbf{99.9}&  40.4 &  25.3 &  \textbf{99.9}&  68.9 &  52.6 &  \textbf{99.7} \\\midrule
   
    \multirow{3}{*}{\rotatebox{90}{$P_A$\hspace{0pt}}}
    & $\mu_1$ &  \textbf{71.2} &  \textbf{63.7} &  80.5&  82.5 &  \textbf{84.1} &  81.0&  83.1 &  \textbf{85.4} &  80.9&  82.7 & \textbf{84.9} &  80.7&  \textbf{77.8} &  95.8 &  65.5 \\
  & $s$ &  55.7 &  39.6 &  94.1&  73.7 &  60.1 &  95.2&  75.2 &  62.1 &  95.3&  74.6 &  61.3 &  95.1&  87.3 &  83.4 &  91.5 \\
  & $\mu_2$ &  37.0 &  22.8 &  98.9&  54.2 &  37.3 &  99.4&  55.4 &  38.4 &  99.2&  54.9 &  37.9 &  99.4&  75.7 &  61.3 &  99.0  \\\bottomrule
    \end{tabular}
\caption{$F_1$-Score, Precision, and Recall on the label noise detection task with different probability sets ($P_T$, $P_S$, $P_A$) and different thresholds provided by Otsu's method ($s$, $\mu_1$, $\mu_2$) to split the dataset into clean and corrupted subsets. Best metrics are presented in bold.}
\label{tab:detection}
\end{table*}

%% file: plot_fig1.tex
\begin{figure*}[t]
\centering
\begin{minipage}[t]{0.32\linewidth}
\centering
\input{plot_average_max_p}
\subcaption{The average maximal probability of the students network prediction. Vertical lines denote the maxima during the training process which can be interpreted as tipping point at which the model start fitting to individual sample details.}
\label{fig:pmax}
\end{minipage}
\hspace{0.01\linewidth}
\begin{minipage}[t]{0.32\linewidth}
\centering
\input{plot_accuracy}
\subcaption{Test accuracy of the teacher network during training. Vertical lines denote the tipping points from Fig. \ref{fig:pmax}. With increasing noise rates, the detected tipping points fit approximately to the maxima of the classification test accuracy.}
\label{fig:acc}
\end{minipage}
\hspace{0.01\linewidth}
\begin{minipage}[t]{0.32\linewidth}
\centering
\input{plot_f1}
\subcaption{Noise detection $F_1$-score on the training set with student maxima from Fig. \ref{fig:pmax}. Dashed lines indicate split by Otsu threshold $s$, while solid lines indicate the split by $\mu_1$. The student maxima fit approximately to the maxima of the detection performance for $\mu_1$ for higher noise rates.}
\label{fig:f1}
\end{minipage}
\caption{Relation between the student's prediction behavior and the classification and noise detection performance. The maximum of the student network's average maximum probability indicates the start of fitting to sample details and thus can be used for early stopping to avoid overfitting. The models for this figure are trained without robust training for 75 epochs to show the standard training behavior.}
\label{fig:accuracy_noise}
\end{figure*}
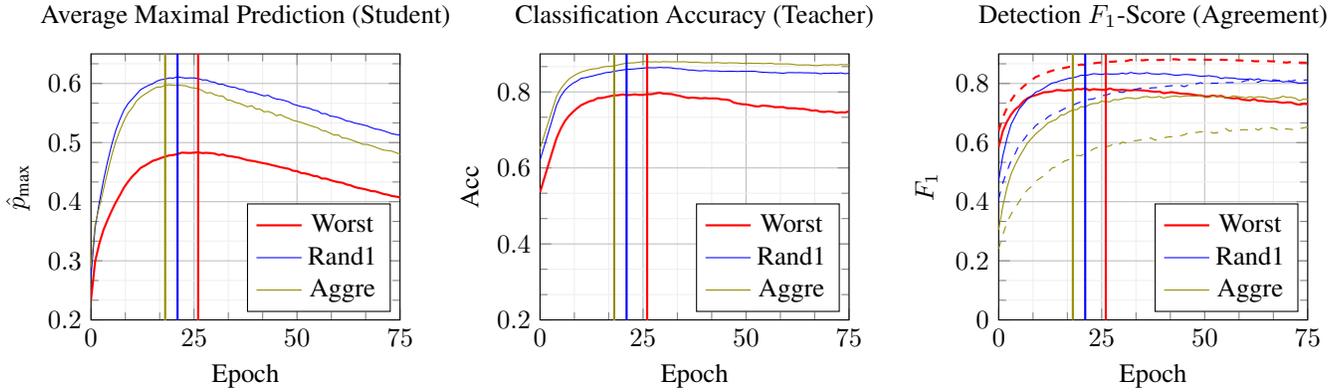

%% file: plot_average_max_p.tex
\begin{tikzpicture}
\begin{axis}[
    title={Average Maximal Prediction (Student)},
    scaled ticks=false, 
    log ticks with fixed point,
    tick label style={/pgf/number format/fixed},
    xmin = 0, xmax = 75,
    ymin = 0.2, ymax = 0.65,
    xtick distance = 25,
    ytick distance = 0.1,
    grid = both,
    minor tick num = 2,
    major grid style = {lightgray},
    minor grid style = {lightgray!25},
    width = 1\linewidth,
    height = 0.9\linewidth,
    xlabel = {Epoch},
    ylabel = {$\hat{p}_\text{max}$},
    legend pos = south east,
    legend style = {
      legend columns=1},
]
\addplot[
    thick,
    red,
] file[skip first] {worse.txt};

\addplot[
    blue,
] file[skip first] {rand.txt};

\addplot[
    olive,
] file[skip first] {aggre.txt};
\addplot +[mark=none,red, thick, solid] coordinates {(26, 0) (26, 0.75)};
\addplot +[mark=none,blue, thick, solid] coordinates {(21, 0) (21, 0.75)};
\addplot +[mark=none,olive, thick, solid] coordinates {(18, 0) (18, 0.75)};
\legend{Worst, Rand1, Aggre}
\end{axis}
\end{tikzpicture}

%% file: plot_accuracy.tex
\begin{tikzpicture}
\begin{axis}[
    title={Classification Accuracy (Teacher)},
    scaled ticks=false, 
    log ticks with fixed point,
    tick label style={/pgf/number format/fixed},
    xmin = 0, xmax = 75,
    ymin = 0.2, ymax = 0.9,
    xtick distance = 25,
    ytick distance = 0.2,
    grid = both,
    minor tick num = 2,
    major grid style = {lightgray},
    minor grid style = {lightgray!25},
    width = 1\linewidth,
    height = 0.9\linewidth,
    xlabel = {Epoch},
    ylabel = {Acc},
    legend pos = south east,
    legend style = {
      legend columns=1},
]
\addplot[
    thick,
    red,
] file[skip first] {worse_acc.txt};

\addplot[
    blue,
] file[skip first] {rand_acc.txt};

\addplot[
    olive,
] file[skip first] {aggre_acc.txt};
\addplot +[mark=none,red, thick, solid] coordinates {(26, 0) (26, 0.9)};
\addplot +[mark=none,blue, thick, solid] coordinates {(21, 0) (21, 0.9)};
\addplot +[mark=none,olive, thick, solid] coordinates {(18, 0) (18, 0.9)};
\legend{Worst, Rand1, Aggre}
\end{axis}
\end{tikzpicture}

%% file: plot_f1.tex
\begin{tikzpicture}
\begin{axis}[
    title={Detection $F_1$-Score (Agreement)},
    scaled ticks=false, 
    log ticks with fixed point,
    tick label style={/pgf/number format/fixed},
    xmin = 0, xmax = 75,
    ymin = 0.0, ymax = 0.90,
    xtick distance = 25,
    ytick distance = 0.2,
    grid = both,
    minor tick num = 2,
    major grid style = {lightgray},
    minor grid style = {lightgray!25},
    width = 1\linewidth,
    height = 0.9\linewidth,
    xlabel = {Epoch},
    ylabel = {$F_1$},
    legend pos = south east,
    legend style = {
      legend columns=1},
]
\addplot[
    thick,
    red,
] table[x=x, y=mu ] {worse_f1.txt};
\addplot[
    blue,
] table[x=x, y=mu ] {rand_f1.txt};
\addplot[
    olive,
] table[x=x, y=mu ] {aggre_f1.txt};
\addplot[
    thick,
    red,
    dashed
] table[x=x, y=s ] {worse_f1.txt};
\addplot[
    blue,
    dashed
] table[x=x, y=s ] {rand_f1.txt};
\addplot[
    olive,
    dashed
] table[x=x, y=s ] {aggre_f1.txt};

\addplot +[mark=none,olive, thick, solid] coordinates {(18, 0) (18, 0.9)};
\addplot +[mark=none,blue, thick, solid] coordinates {(21, 0) (21, 0.9)};
\addplot +[mark=none,red, thick, solid] coordinates {(26, 0) (26, 0.9)};
\legend{Worst, Rand1, Aggre}
\end{axis}
\end{tikzpicture}

%% file: plot_accuracy_worse.tex
\begin{figure}[t]
\centering

\begin{tikzpicture}
\begin{axis}[
    scaled ticks=false, 
    log ticks with fixed point,
    tick label style={/pgf/number format/fixed},
    xmin = 0, xmax = 75,
    ymin = 0.45, ymax = 0.85,
    xtick distance = 25,
    ytick distance = 0.2,
    grid = both,
    minor tick num = 2,
    major grid style = {lightgray},
    minor grid style = {lightgray!25},
    width = 0.9\linewidth,
    height = 0.6\linewidth,
    xlabel = {Epoch},
    ylabel = {Acc},
    legend pos = south east,
    legend style = {
      legend columns=1},
]
\addplot[
    thick,
    black,
] table[x=x, y=teacher ] {worse_acc_all.txt};

\addplot[
    blue,
] table[x=x, y=student ] {worse_acc_all.txt};

\addplot[
    green,
] table[x=x, y=agreement ] {worse_acc_all.txt};
\addplot +[mark=none,black, thick, solid] coordinates {(26, 0) (26, 0.9)};
\legend{$P_T$ (Teacher), $P_S$ (Student), $P_A$ (Agreement)}
\end{axis}
\end{tikzpicture}

\caption{Test accuracy w.r.t.~training epoch based on the predicted probabilities of the teacher, student, and the proposed combined agreement. The agreement probability combines the strengths and outperforms the teacher and student probability. Note that we train the framework for 75 epochs and without robust optimization (Sec.~\ref{sec:robust_optimization}).}
\label{fig:accuracy_worse_all}
\end{figure}
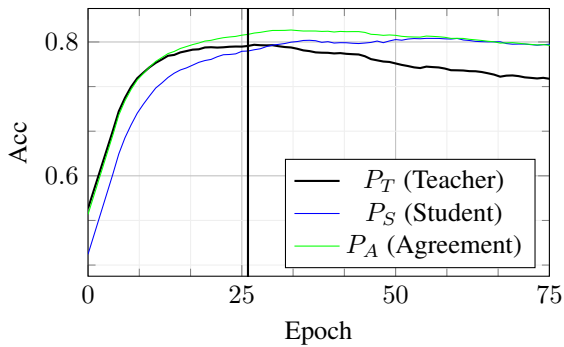

%% file: 5_conclusion.tex
\section{Conclusion}
This paper introduces \textit{Blind Knowledge Distillation} that is able to transfer simple and general image patterns that are not based on individual image details. We show that our framework is able to identify the tipping point from fitting to simple but general image patterns to fitting to image details and use it for early stopping in standard classification frameworks and furthermore to estimate the likelihood of samples in the training data of being clean or corrupted.

Our method performs on par with state-of-the-art methods that are not extended with high performance semi-supervised training strategies. Compared to them, we do not rely on manually predefined warm-up phases and adapt it online during training. However, the intention of this paper is to provide new insights about general learning behavior rather than to tune our method with known strategies. We hope that \textit{Blind Knowledge Distillation} helps researchers to improve the handling of under- and overfitting. 

%% file: 6_acknowledgements.tex
\section*{Acknowledgments}
This work was supported by the Federal Ministry of Education and Research (BMBF), Germany under the project LeibnizKILabor (grant no. 01DD20003) and the Deutsche Forschungsgemeinschaft  (DFG)  under  Germany’s  Excellence  Strategy  within  the  Cluster of Excellence PhoenixD (EXC 2122).